\documentclass[10pt,twocolumn,letterpaper]{article}

\usepackage{iccv}
\usepackage{times}
\usepackage{epsfig}
\usepackage{graphicx}
\usepackage{amsmath}
\usepackage{amssymb}
\usepackage{booktabs}
\usepackage{algorithm}
\usepackage[algo2e]{algorithm2e} 
\usepackage{algorithmic}
\usepackage[breaklinks=true,bookmarks=false]{hyperref}
\usepackage[accsupp]{axessibility}
\iccvfinalcopy
\def\iccvPaperID{6000} 

\ificcvfinal\fi

\begin{document}

%%%%%%%%% TITLE
\title{Towards Better Explanations of Class Activation Mapping}

\author{Hyungsik Jung \qquad Youngrock Oh \\ 
Samsung SDS \\
{\tt\small \{hs89.jung,y52.oh\}@samsung.com}
}

\maketitle
\thispagestyle{empty}

%%%%%%%%% ABSTRACT
\begin{abstract}
   Increasing demands for understanding the internal behavior of convolutional neural networks (CNNs) have led to remarkable improvements in explanation methods. Particularly, several class activation mapping (CAM) based methods, which generate visual explanation maps by a linear combination of activation maps from CNNs, have been proposed. However, the majority of the methods lack a clear theoretical basis on how they assign the coefficients of the linear combination. In this paper, we revisit the intrinsic linearity of CAM with respect to the activation maps; we construct an explanation model of CNN as a linear function of binary variables that denote the existence of the corresponding activation maps. With this approach, the explanation model can be determined by additive feature attribution methods in an analytic manner. We then demonstrate the adequacy of SHAP values, which is a unique solution for the explanation model with a set of desirable properties, as the coefficients of CAM. Since the exact SHAP values are unattainable, we introduce an efficient approximation method, LIFT-CAM, based on DeepLIFT. Our proposed LIFT-CAM can estimate the SHAP values of the activation maps with high speed and accuracy. Furthermore, it greatly outperforms other previous CAM-based methods in both qualitative and quantitative aspects.
\end{abstract}

%%%%%%%%% BODY TEXT
\section{Introduction}

Recently, convolutional neural networks (CNNs) have achieved excellent performance in various real-world vision tasks. However, it is difficult to explain their predictions due to a lack of understanding of their internal behavior. To grasp why a model makes a certain decision, numerous \emph{saliency methods} have been proposed. The methods generate \emph{visual explanation maps} that represent pixel-level importances for which regions in an input image are responsible for the model's decision and which are not. Towards better comprehension of CNNs, \emph{class activation mapping (CAM)} based methods, which utilize the responses of a convolutional layer for explanations, have been widely used.\par 

CAM-based methods \cite{grad_cam_plus, ablation_cam, xgrad_cam, grad_cam, score_cam, base_cam} (abbreviated as CAMs in the remainder of this paper) \emph{linearly} combine \emph{activation maps} to produce visual explanation maps. Since the activation maps are fixed for a given input image and a model pair, the coefficients of a linear combination govern the performance of the methods. Therefore, it is critical to design a reasonable method of determining the coefficients. However, the majority of CAMs rely on heuristic conjectures for coefficient assignment without a clear theoretical basis. Specifically, the underlying linearity of CAM w.r.t. the activation maps is not fully taken into account. In addition, they do not set rigorous standards of which properties are expected to be satisfied in a good explanation model.\par

In this work, we leverage the linearity of CAM to analytically determine the coefficients beyond heuristics. Focusing on the fact that CAM defines an explanation map using a linear combination of activation maps, we formulate an explanation model as a linear function of the binary variables denoting the existence of the associated activation maps. Under this scheme, each activation map can be seen as an individual feature in \emph{additive feature attribution methods} \cite{lrp, SHAP, lime, deep_lift}. Notably, \emph{SHapley Additive exPlanations (SHAP)} \cite{SHAP} provides SHAP values as a unified measure of feature importance that satisfies three desirable properties (described in Sec. \ref{sec:2.2}). Thus, the coefficients can be determined by the SHAP values of the corresponding activation maps. However, the exact SHAP values are not computable. To solve this, we propose a novel saliency method using \emph{Deep Learning Important FeaTures (DeepLIFT)} \cite{deep_lift}, called \emph{LIFT-CAM}, which efficiently approximates the SHAP values of the activation maps. Our contributions are summarized as follows:

\begin{itemize}
    \item We propose a novel framework of determining a plausible visual explanation map of CAM, by reframing the problem as determining a reliable solution for the explanation model using additive feature attribution methods. The recent Ablation-CAM \cite{ablation_cam} can be reinterpreted by this framework.
    \item We formulate the SHAP values of the activation maps as a unified solution for the proposed framework and verify their benefits in terms of generating faithful visual explanations.
    \item We introduce a new saliency method, LIFT-CAM, based on DeepLIFT. It effectively estimates the SHAP values of the activation maps with a single backward propagation and outperforms the other previous CAMs qualitatively and quantitatively.
\end{itemize}

%-------------------------------------------------------------------------
\section{Related Work}
\subsection{Class Activation Mapping}\label{sec:2.1}
\noindent \textbf{Visual explanation map.} Let $f$ be an original prediction model and $c$ denote a target class of interest. CAMs \cite{grad_cam_plus, ablation_cam, xgrad_cam, grad_cam, score_cam, base_cam} aim to explain the target output of the model for a specific input image $x$ (i.e., $f^{c}(x)$) through the visual explanation map, which can be generated by:
\begin{equation}\label{eqn:general_CAM}
L_{\text{CAM}}^{c}(A) = \text{ReLU}(\sum\limits_{k=1}^{N_{l}}\alpha_{k}A_{k})
\end{equation}
with $A=f^{[l]}(x)$, where $f^{[l]}(\cdot)$ denotes the output of the $l$-th layer\footnote{Conventionally, the last convolutional layer is used for the layer $l$ because it is expected to provide the best compromise between high-level semantics and spatial information \cite{grad_cam}.}. $A_{k}$ is a $k$-th activation map of $A$ and $\alpha_{k}$ is the coefficient (i.e., the importance) of $A_{k}$, respectively. $N_{l}$ indicates the number of the activation maps of the $l$-th layer. This concept of linearly combining activation maps was firstly proposed by \cite{base_cam}, leading to its variants.\par

\noindent \textbf{Previous methods.} Grad-CAM \cite{grad_cam} decides the coefficient of a specific activation map by averaging the gradients over all activation neurons in that map. Grad-CAM++ \cite{grad_cam_plus}, which is a modified version of Grad-CAM, focuses on positive influences of neurons considering higher-order derivatives. However, the gradients of deep neural networks tend to diminish due to the gradient saturation problem. Hence, using unmodified raw gradients induces failure of localization for relevant regions. \par
To overcome this limitation, gradient-free CAMs have been proposed. Score-CAM \cite{score_cam} overlaps normalized activation maps to an input image and makes predictions to acquire the coefficients. Ablation-CAM \cite{ablation_cam} defines a coefficient as the fraction of decline in the target output when the associated activation map is removed. They are free from the saturation issue, but time-consuming because they require $N_{l}$ forward propagations to acquire the coefficients. \par
All the methods described above determine their coefficients in a heuristic way. XGrad-CAM \cite{xgrad_cam} addresses this issue by suggesting two axioms. The authors derived the coefficients that satisfy the axioms as much as possible. However, their derivation is demonstrated only for ReLU-CNNs. \par

\subsection{SHapley Additive exPlanations}\label{sec:2.2}
\noindent \textbf{Additive feature attribution method.} SHAP \cite{SHAP} is a unified explanation framework for additive feature attribution methods. The methods follow:
\begin{equation}\label{eqn:afam}
g(z^{'}) = \phi_{0}+\sum\limits_{i=1}^{M}\phi_{i}z^{'}_{i}
\end{equation}
where $g$ is an explanation model of an original prediction model $f$ for a specific input $x$ and a target class $c$. $M$ is the number of input features and $z^{'} \in \{0,1\}^{M}$ indicates a binary vector in which each entry represents the existence of the corresponding original input feature; $1$ for \emph{presence} and $0$ for \emph{absence}. $\phi_{i}$ denotes an importance of the $i$-th feature and $\phi_{0}$ is a baseline explanation. The methods are designed to ensure $g(z^{'})\approx f^{c}(h_{x}(z^{'}))$ whenever $z^{'} \approx x^{'}$, with a mapping function $h_{x}$ that satisfies $x=h_{x}(x^{'})$. While several existing attribution methods \cite{lrp, SHAP, lime, deep_lift} match Eq. \eqref{eqn:afam}, only one explanation model satisfies three desirable properties: \emph{local accuracy}, \emph{missingness}, and \emph{consistency} \cite{SHAP}. \par

\noindent \textbf{SHAP values.} A feature attribution of the explanation model which obeys Eq. \eqref{eqn:afam} while adhering to the above three properties is defined as SHAP values \cite{SHAP} and can be formulated by:
\begin{equation}\label{eqn:SHAP}
\phi_{i} = \sum\limits_{z^{'}\subset x^{'}}\frac{(M-\lvert z^{'} \rvert)!(\lvert z^{'} \rvert-1)!}{M!}[f^{c}(h_{x}(z^{'}))-f^{c}(h_{x}(z^{'}\text{\textbackslash $i$}))]
\end{equation}
where $\lvert z^{'} \rvert$ denotes the number of non-zero entries in $z^{'}$ and $z^{'} \subset x^{'}$ indicates all $z^{'}$ vectors, where the non-zero entries are a subset of the non-zero entries in $x^{'}$. In addition, $z^{'}\text{\textbackslash} i$ means setting $z_{i}^{'}=0$. This definition of the SHAP values intimately aligns with the classic Shapley values \cite{shapley}.

\subsection{Deep Learning Important FeaTures}
DeepLIFT \cite{deep_lift} focuses on the difference between an \emph{original} activation and a \emph{reference} activation. It propagates the difference through a network to assign the contribution score to each input feature by linearizing non-linear components in the network. Through this technique, the gradient saturation problem is alleviated. \par
Let $o$ represent the output of the target neuron and $x = (x_{1}, \ldots, x_{n})$ be inputs whose reference values are $r = (r_{1}, \ldots, r_{n})$.
The contribution score of the $i$-th input feature $C_{\Delta x_{i} \Delta o}$ quantifies the influence of $\Delta x_{i}=x_{i}-r_{i}$ on $\Delta o = f^{c}(x)-f^{c}(r)$. In addition, DeepLIFT satisfies the \emph{summation-to-delta} property as below:
\begin{equation}\label{eqn:summation_to_delta}
    \sum\limits_{i=1}^{n}C_{\Delta x_{i} \Delta o} = \Delta o.
\end{equation}
Note that if we set $C_{\Delta x_{i}\Delta o}=\phi_{i}$ and $f^{c}(r)=\phi_{0}$, then Eq. \eqref{eqn:summation_to_delta} matches Eq. \eqref{eqn:afam}. Therefore, DeepLIFT is also an additive feature attribution method. It \emph{approximates} the SHAP values efficiently, satisfying the local accuracy and missingness \cite{SHAP}.
%of which the contribution scores are among the solutions for Eq. \eqref{eqn:afam}

%--------------------------------------------------------------------------
\section{Methodology}
In this section, we clarify the problem formulation of CAM and suggest an approach to solve it analytically. First, we propose a framework that defines a linear explanation model and determines the coefficients of CAM based on the model. Then, we formulate the SHAP values of the activation maps as a unified solution for the framework. Finally, we introduce a fast approximation method for the SHAP values of the activation maps: LIFT-CAM. \par

\subsection{Problem formulation of CAM}
As identified in Eq. \eqref{eqn:general_CAM}, CAM produces a visual explanation map $L^c_{\text{CAM}}$ linearly w.r.t. the activation maps $A_{1},\ldots,A_{N_{l}}$ except for ReLU, which is applied for the purpose of only considering the positive influence on the target class $c$. In addition, the complete activation maps $A$ does not change for a given pair of the model $f$ and the input image $x$. Thus, the quality of $L^c_{\text{CAM}}$ is controlled by the coefficients $\alpha=(\alpha_{1},\ldots,\alpha_{N_{l}})$, which represent the importance scores of the associated activation maps. In sum, the purpose of CAM is to find $\alpha$ for a linear combination in order to generate $L^c_{\text{CAM}}$, which can reliably explain the target output $f^{c}(x)$.

\subsection{Proposed framework}
\emph{How can we acquire the desirable $\alpha$ in an analytic way?} To this end, we first consider each activation map as an individual feature (i.e., we have $N_{l}$ features) and define a binary vector $a^{'} \in \{0,1\}^{N_{l}}$ of the features. In the vector, an entry $a^{'}_{k}$ of $1$ indicates that the corresponding $A_{k}$ maintains its original activation values, and $0$ means that it loses the values.\par
Next, we specify an explanation model $g_{\text{CAM}}$ to interpret $f^c(x)$. Since the explanation map of CAM $L^c_{\text{CAM}}$ is linear w.r.t. the activation maps $A_{1},\ldots,A_{N_{l}}$ by definition, it is reasonable to assume that the explanation model $g_{\text{CAM}}$ is also linear w.r.t. the binary variables of the activation maps $a^{'}_{1},\ldots,a^{'}_{N_{l}}$ as follows:
\begin{equation}\label{eqn:afam_act}
g_{\text{CAM}}(a^{'}) = \alpha_{0}+\sum\limits_{k=1}^{N_{l}}\alpha_{k}a^{'}_{k}.
\end{equation}
Under this assumption, the problem of determining $\alpha$ in Eq. \ref{eqn:general_CAM} can be reformulated into the problem of determining $g_{\text{CAM}}$ that follows Eq. \ref{eqn:afam_act}.\par
\begin{figure}[t]
\begin{center}
   \includegraphics[width=3.35in]{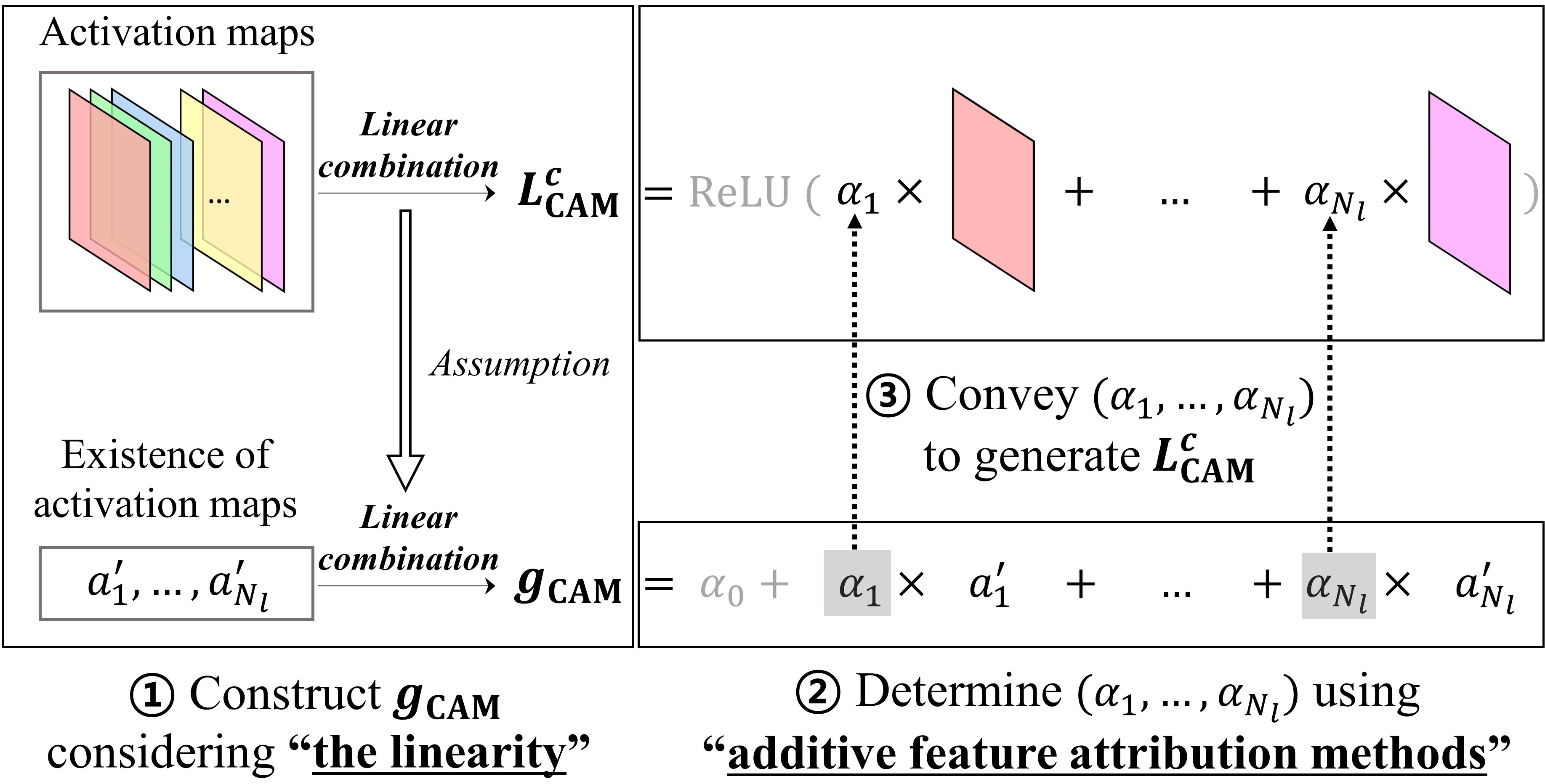}
\end{center}
   \caption{Proposed framework for determining the coefficients of CAM. First, we build a linear explanation model. Next, we determine the importance scores of the activation maps by optimizing the explanation model, using additive feature attribution methods. Last, we use the scores as the coefficients of CAM.}
\label{fig:main_concept}
\end{figure}

Eq. \ref{eqn:afam_act} matches Eq. \ref{eqn:afam} exactly. Furthermore, each individual feature of Eq. \ref{eqn:afam_act} (i.e., each activation map) is expected to represent distinct high-level semantic information. Therefore, in this work, we determine $g_{\text{CAM}}$ using additive feature attribution methods (\cite{SHAP} in Sec. \ref{sec:3.3}, \cite{deep_lift} in Sec. \ref{sec:3.4}, \cite{lrp} and \cite{lime} in Supplementary Material). Once we obtain $\alpha$ on the basis of $g_{\text{CAM}}$, we can use the values to generate $L^c_{\text{CAM}}$. Figure \ref{fig:main_concept} shows our proposed framework which is described in this section.\par
% Therefore, the SHAP values can be adopted as a \emph{unified} solution for $\{\alpha_{k}\}_{k\in\{1,\ldots,N_{l}\}}$ (see the supplementary materials for comparison with LIME \cite{lime})
%This linear explanation model can be solved by additive feature attribution methods, while matching Eq. \eqref{eqn:afam} exactly. Therefore, in this work, we reformulate the problem of determining $\alpha$ to solving Eq. \eqref{eqn:afam_act} by using several existing additive feature attribution methods 

\begin{figure*}
\begin{center}
    \includegraphics[width=6.5in]{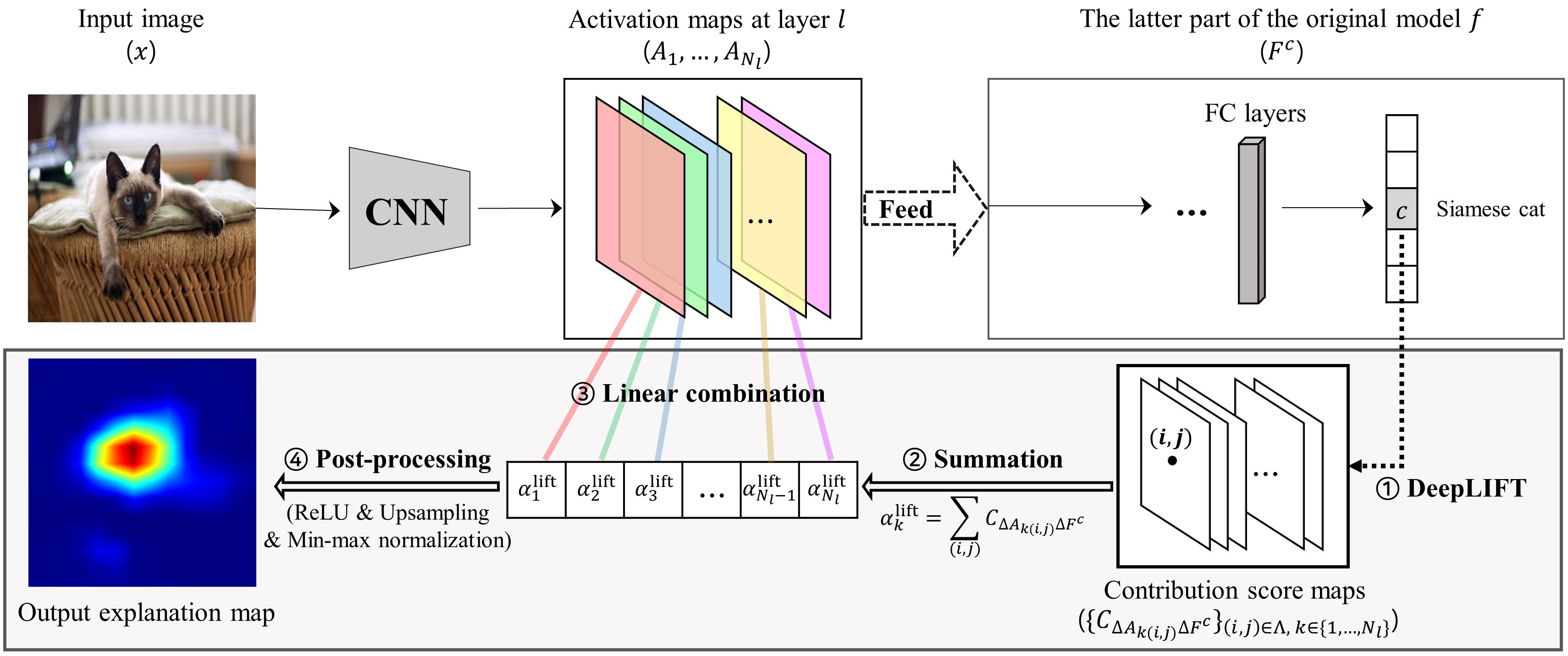}
\end{center}
   \caption{Overview of LIFT-CAM. First, we apply DeepLIFT from the target output up to the activation maps and acquire the contribution score maps, in which each pixel represents $C_{\Delta A_{k}{_{(i,j)}} \Delta F^{c}}$. Next, we quantify the importance of each activation map by summing all the contribution scores of itself. Then, we perform a linear combination of $(\alpha_{1}^{\text{lift}},\ldots,\alpha_{N_{l}}^{\text{lift}})$ and $A_{1},\ldots,A_{N_{l}}$. Finally, we rectify the resulting map, upsample the map to the original image dimension, and normalize the map using the min-max normalization function.}
\label{fig:lift_cam}
\end{figure*}

\subsection{SHAP values of activation maps}\label{sec:3.3}
SHAP \cite{SHAP} is a model-agnostic method and the local accuracy, missingness, and consistency \cite{SHAP} are still desirable in Eq. \ref{eqn:afam_act}. Accordingly, we adopt the SHAP values of the activation maps as a unified solution for our framework. Let $F$ be a latter part of the original model $f$, from layer $l+1$ to layer $L-1$\footnote{It represents the logit layer which precedes the softmax layer.}, where $L$ represents the total number of layers in $f$. Namely, we have $F(A) = f^{[L-1]}(x)$. Additionally, we define a mapping function $h_{A}$ that converts $a^{'}$ into the embedding space of $A$; it satisfies $A=h_{A}(A^{'})$, where $A^{'}$ is a vector of ones. Specifically, $a^{'}_{k}=1$ is mapped to $A_{k}$ and $a^{'}_{k}=0$ to \textbf{0}, which has the same dimension as $A_{k}$. Note that this is reasonable because $A_{k}$ exerts no influence on $L^{c}_{\text{CAM}}$ when it has values of $0$ for all activation neurons in Eq. \eqref{eqn:general_CAM}. \par

Now, the SHAP values of the activation maps w.r.t. the class $c$ are formulated by:
\begin{equation}\label{eqn:SHAP_act}
\alpha_{k}^{\text{shap}} = \sum\limits_{a^{'}\subset A^{'}}\frac{(N_{l}-\lvert a^{'} \rvert)!(\lvert a^{'} \rvert-1)!}{N_{l}!}[F^{c}(h_{A}(a^{'}))-F^{c}(h_{A}(a^{'}\text{\textbackslash $k$}))]
\end{equation}
where $\alpha_{k}^{\text{shap}}$ is the SHAP value of $A_{k}$ and $F^c$ denotes the target output of $F$. The above equation implies that $\alpha_{k}^{\text{shap}}$ can be obtained by averaging marginal prediction differences between \emph{presence} and \emph{absence} of $A_{k}$ across $\Pi_{\text{total}}$ that denotes a set of all possible feature orderings of $\{1, \ldots, N_l\}$. \par
To reduce the computational burden, we suggest to use a subset $\Pi \subsetneq \Pi_{\text{total}}$ instead of $\Pi_{\text{total}}$ to estimate $\alpha^{\text{shap}}=(\alpha^{\text{shap}}_{1},\ldots,\alpha^{\text{shap}}_{N_{l}})$, as described in Algorithm \ref{algorithm:shap}. We refer to the algorithm of using $\lvert \Pi \rvert$ orderings as SHAP-CAM$_{\lvert \Pi \rvert}$ throughout the paper.
The higher $\lvert \Pi \rvert$, the $\alpha$ from SHAP-CAM$_{\lvert \Pi \rvert}$ converges to $\alpha^{\text{shap}}$ by the law of large numbers. We validate the benefits of these SHAP attributions in terms of the faithfulness of $L^{c}_{\text{CAM}}$ in Sec. \ref{sec:4.1}. Analyses for acknowledged approximation methods for SHAP values, DeepSHAP \cite{SHAP} and KernelSHAP \cite{SHAP}, are provided in Supplementary Material.

\begin{algorithm}[H]
%\SetAlgoLined
\SetKwInOut{Input}{Input}
\SetKwInOut{Output}{Output}
\SetKw{Initialize}{Initialize:}
\Input{$F$, $c$,
$h_{A}$, and a subset $\Pi \subsetneq \Pi_{\text{total}}$}
\Output{$\alpha=(\alpha_{1},\ldots,\alpha_{N_{l}})$}
\Initialize{$\alpha \xleftarrow[]{}\mathbf{0}$}

\For{\emph{each ordering} $\pi$ \emph{in} $\Pi$}{
$a^{'} \xleftarrow[]{}\mathbf{0}$

  \For{$i=1,...,N_{l}$}{
  $a^{'}_{\pi(i)} \xleftarrow[]{} 1$
  
  $\alpha_{\pi(i)} \xleftarrow[]{} \alpha_{\pi(i)} + F^{c}(h_{A}(a^{'})) - F^{c}(h_{A}(a^{'}\text{\textbackslash $\pi(i)$}))$
  }
 }
 $\alpha \xleftarrow[]{} \alpha / \lvert \Pi \rvert$ 
 \caption{SHAP-CAM$_{\lvert \Pi \rvert}$}
 \label{algorithm:shap}
\end{algorithm}

\subsection{Efficient approximation: LIFT-CAM}\label{sec:3.4}
Through the experiment in Sec. \ref{sec:4.1}, we demonstrate that a faithful $L^{c}_{\text{CAM}}$ can be achieved by $\alpha^{\text{shap}}$. However, calculating the exact $\alpha^{\text{shap}}$ is almost impossible. Therefore, we need to consider an approximation approach. In this study, we propose a novel method, LIFT-CAM, that efficiently approximates $\alpha^{\text{shap}}$ using DeepLIFT\footnote{DeepLIFT-Rescale is used for approximation because the method can be easily implemented by overriding gradient operators. This convenience enables LIFT-CAM to be easily applied to a large variety of tasks.} \cite{deep_lift}.\par
First, we calculate the contribution score for every activation neuron at layer $l$ using DeepLIFT through a single backward pass. Considering the summation-to-delta property of DeepLIFT, we define the contribution score of a specific activation map as the summation of the contribution scores of all neurons in that activation map, as follows:
\begin{equation}
    \alpha_{k}^{\text{lift}} = C_{\Delta A_{k} \Delta F^{c}} = \sum\limits_{(i,j)\in \Lambda}C_{\Delta A_{k}{_{(i,j)}} \Delta F^{c}}
\end{equation}
where $\Lambda = \{1, \dots, H\} \times \{1, \dots, W\}$ is a discrete activation dimension and $A_{k}{_{(i,j)}}$ is an activation value at the $(i,j)$ location of $A_{k}$. Note that $\Delta$ denotes the difference-from-reference and the reference values (i.e., the values corresponding to the \emph{absent} feature) of all activation neurons are set to 0, aligning with SHAP. By this definition, $\alpha^{\text{lift}}=(\alpha^{\text{lift}}_{1},\ldots,\alpha^{\text{lift}}_{N_{l}})$ becomes a reliable solution for Eq. \eqref{eqn:afam_act} while satisfying the local accuracy\footnote{$\sum^{N_{l}}_{k=1} \alpha_{k}^{\text{shap}} = F^c(A)-F^c(\mathbf{0}).$} of SHAP as below:
\begin{equation}
    \sum^{N_{l}}_{k=1} \alpha_{k}^{\text{lift}} = F^c(A)-F^c(\mathbf{0}).
\end{equation}
Consequently, LIFT-CAM can estimate $\alpha^{\text{shap}}$ with a single backward pass while alleviating the gradient saturation problem \cite{deep_lift}. Figure \ref{fig:lift_cam} shows an overview of our proposed LIFT-CAM. Additionally, the following rationale motivates us to employ DeepLIFT for this problem.\par
DeepLIFT linearizes non-linear components to estimate the SHAP values \cite{SHAP}. Therefore, DeepLIFT attributions tend to deviate from the true SHAP values when passed through many overlapping non-linear layers during back-propagation (see Supplementary Material for details). However, for CAM, only the non-linearities in $F$ matter. Since CAM usually uses the outputs of the last convolutional layer as its activation maps, almost all $F$ of state-of-the-art architectures contain few non-linearities (e.g., the VGG family), or are even fully linear (e.g., the ResNet family). Thus, the SHAP values of the activation maps can be approximated with high precision by LIFT-CAM. Particularly, we can acquire the exact SHAP values using LIFT-CAM (i.e., $\alpha^{\text{lift}}=\alpha^{\text{shap}}$) for the architectures of linear $F$. The proof of this statement is provided in Supplementary Material.

\subsection{Rethinking Ablation-CAM}
The recently proposed Ablation-CAM \cite{ablation_cam} can be reinterpreted by our framework. Ablation-CAM defines the coefficients as below:
\begin{equation}
    \alpha_{k}^{\text{ablation}}=\frac{F^{c}(h_{A}(A^{'}))-F^{c}(h_{A}(A^{'} \text{\textbackslash}k))}{F^{c}(h_{A}(A^{'}))}.
\end{equation}
Since Ablation-CAM uses this specific marginal difference as the coefficient, it can be deemed as another approximation method for $\alpha^{\text{shap}}$. However, Ablation-CAM is computationally expensive requiring $N_{l}$ forward simulations. In addition, the method does not satisfy the local accuracy of SHAP (i.e., $\sum^{N_{l}}_{k=1} \alpha_{k}^{\text{ablation}} \neq F^c(A)-F^c(\mathbf{0})$). This mismatch leads to less precise approximations compared to LIFT-CAM, resulting in less reliable explanations. \\
%\noindent \textbf{LRP-CAM.} Layer-wise relevance propagation (LRP) \cite{lrp} is an additive feature attribution method that conserves the sum of relevance scores between layers, like LIFT-CAM. Therefore, we can define LRP-CAM to have:
%\begin{equation}
%    \alpha^{\text{lrp}}_{k}=\sum_{(i,j)\in \Lambda}R(A_{k}{_{(i,j)}})
%\end{equation}
%where $R(A_{k}{_{(i,j)}})$ is the relevance score of $A_{k}{_{(i,j)}}$ w.r.t. $F^{c}(A)$. These LRP attributions $\alpha^{\text{lrp}}=(\alpha^{\text{lrp}}_{1},\ldots,\alpha^{\text{lrp}}_{N_{l}})$ estimate $\alpha^{\text{shap}}$ as a solution for Eq. \eqref{eqn:afam_act}. LRP-CAM needs only a single backward propagation to obtain $\alpha^{\text{lrp}}$. \par

\section{Experiments and Results}
We now describe our experiments and show the results. In Sec. \ref{sec:4.1}, we first validate the superiority of $\alpha^{\text{shap}}$ by evaluating the faithfulness of $L^{c}_{\text{CAM}}$ generated by SHAP-CAM. Then, we demonstrate how closely LIFT-CAM can estimate $\alpha^{\text{shap}}$ in Sec. \ref{sec:4.2}. These two experiments provide justification to opt for LIFT-CAM as a responsible method of determining $\alpha$ of CAM. We then evaluate the performance of LIFT-CAM on the object recognition task in the context of image classification, comparing it with the other CAMs: Grad-CAM, Grad-CAM++, XGrad-CAM, Score-CAM, and Ablation-CAM in Sec. \ref{sec:4.3}. Finally, we apply LIFT-CAM to the visual question answering (VQA) task in Sec. \ref{sec:4.4} to check the scalability of the method.\par
For all experiments except VQA, we employ the public classification datasets: ImageNet \cite{imagenet} (ILSVRC 2012 validation set), PASCAL VOC \cite{voc} (2007 test set), and MS COCO \cite{coco} (2014 validation set). In addition, the VGG16 network trained on each dataset is analyzed for the experiments (see Supplementary Material for the experiments of the ResNet50). We refer to the pretrained models from the torchvision\footnote{https://github.com/pytorch/vision/blob/master/torchvision} package for ImageNet and the TorchRay package\footnote{https://github.com/facebookresearch/TorchRay} for VOC and COCO. For VQA, we use the fundamental architecture\footnote{https://github.com/tbmoon/basic\_vqa} proposed by \cite{vqa} and the VQA v2.0 dataset \cite{vqa2}.

\subsection{Validation of SHAP values}\label{sec:4.1}

\begin{table*}[t]
\begin{center}
\begin{tabular}{l c c c c c c c c c}
\toprule
     & \multicolumn{3}{c}{Increase in Confidence (\%)} & \multicolumn{3}{c}{Average Drop (\%)} & \multicolumn{3}{c}{Average Drop in Deletion (\%)} \\ 
\cmidrule(lr){2-4} \cmidrule(lr){5-7} \cmidrule(lr){8-10}
    & \emph{ImageNet} & \emph{VOC} & \emph{COCO} & \emph{ImageNet} & \emph{VOC} & \emph{COCO} & \emph{ImageNet} & \emph{VOC} & \emph{COCO} \\
\midrule
    $^{*}$SHAP-CAM$_{1}$ & 25.9 & 37.4 & 35.2 & 28.16 & 22.93 & 23.98 & 32.64 & 17.35 & 24.07\\
    $^{*}$SHAP-CAM$_{10}$ & 26.2 &  42.7 & 40.1 & 27.54 & 17.53 & 19.07 & 32.99 & 19.95 & 27.50 \\
    $^{*}$SHAP-CAM$_{100}$ & \underline{\textbf{26.4}} & \underline{\textbf{43.6}} & \underline{\textbf{41.4}} & \underline{\textbf{27.48}} & \underline{\textbf{16.71}} & \underline{\textbf{18.27}} & \underline{\textbf{33.03}} & \underline{\textbf{20.64}} & \underline{\textbf{27.65}} \\
\bottomrule
\end{tabular}
\end{center}
\caption{Faithfulness evaluation on the object recognition task for SHAP-CAM$_{1}$, SHAP-CAM$_{10}$, and SHAP-CAM$_{100}$. The symbol * denotes averaging for 10 runs. We analyze 1,000 randomly selected images for each dataset. Higher is better for the IC and ADD. Lower is better for the AD.}
\label{table:proof_shap}
\end{table*}

\noindent \textbf{Faithfulness evaluation metrics.} Intuitively, an explanation image w.r.t. the target class $c$ can be generated using an original image $x$ and a related visual explanation map $L^c_{\text{CAM}}$ as below:
\begin{equation}\label{eqn:perturbed_image}
e^{c} = s(u(L^c_{\text{CAM}})) \circ x 
\end{equation}
where $u(\cdot)$ indicates the upsampling operation into the original image dimension and $s(\cdot)$ denotes the min-max normalization function. The operator $\circ$ refers to the Hadamard product. Hence, $e^{c}$ preserves the information of $x$ only in the region which $L^c_{\text{CAM}}$ considers important. \par
In general, $L^c_{\text{CAM}}$ is expected to recognize the regions which contribute the most to the model's decision. Thus, we can evaluate the faithfulness of $L^c_{\text{CAM}}$ on the object recognition task via the two metrics proposed by \cite{grad_cam_plus}: Increase in Confidence (IC) and Average Drop (AD), which are defined as:
\begin{equation}\label{eqn:IC}
    \text{IC} = \frac{1}{N}\sum_{i=1}^{N}\mathbf{1}_{[Y^{c}_{i} < O^{c}_{i}]}\times100,
\end{equation}
\begin{equation}\label{eqn:AD}
    \text{AD} = \frac{1}{N}\sum_{i=1}^{N}\frac{\text{max}(0,Y^{c}_{i}-O^{c}_{i})}{Y^{c}_{i}}\times100,
\end{equation}
where $Y^{c}_{i}$ and $O^{c}_{i}$ are the model's softmax outputs of an $i$-th input image $x_{i}$ and the associated explanation image $e^{c}_{i}$, respectively. $N$ denotes the number of images and $\mathbf{1}_{[\cdot]}$ is an indicator function. Higher is better for the IC and lower is better for the AD.\par
However, both IC and AD evaluate the performance of the explanations via the \emph{preservation} perspective; the region which is considered to be influential is maintained. We can also evaluate the performance through the opposite perspective (i.e., \emph{deletion}); if we mute the region which is responsible for the target output, the softmax probability is expected to drop significantly. From this viewpoint, we suggest the Average Drop in Deletion (ADD) which can be defined as below:
\begin{equation}
    \text{ADD}=\frac{1}{N}\sum_{i=1}^{N}\frac{(Y^{c}_{i}-D^{c}_{i})}{Y^{c}_{i}}\times100
\end{equation}
where $D^{c}$ is the softmax output of the inverted explanation image $e^{c}_{\text{inv}}=(\mathbf{1}-s(u(L^c_{\text{CAM}}))) \circ x$. Higher is better for this metric.\par

%generated by Algorithm \ref{algorithm:shap} of which the weight coefficients are calculated by averaging the marginal contributions from 1, 10, and 100 randomly sampled orderings. We now evaluate the visual explanation maps of SHAP-CAM$_{1}$, SHAP-CAM$_{10}$, and SHAP-CAM$_{100}$. Table \ref{table:proof_shap} shows IC, AD, and ADD of them. We analyze 1,000 randomly selected images for each dataset. Furthermore, each case is averaged over 10 simulations for fair comparison.
\noindent \textbf{Faithfulness evaluation.} Table \ref{table:proof_shap} presents the comparative results of the IC, AD, and ADD between SHAP-CAM$_{1}$, SHAP-CAM$_{10}$, and SHAP-CAM$_{100}$. Each case is averaged for 10 simulations. We discover two important implications from Table \ref{table:proof_shap}. First, as ${\lvert \Pi \rvert}$ increases, the IC and ADD increase and the AD decreases, showing performance improvement. This result indicates that the closer the importances of the activation maps are to $\alpha^{\text{shap}}$, the more effectively the distinguishable region of the target object is found. Second, even compared to the other CAMs (see Table \ref{table:comp_cam}), SHAP-CAM$_{100}$ shows the best performances for all cases. It reveals the adequacy of $\alpha^{\text{shap}}$ as the coefficients of CAM. However, this approach of averaging the marginal contributions of multiple orderings is impractical due to the significant computational burden. Therefore, we propose a cleverer approximation method: LIFT-CAM.\par

%Since the exact SHAP values are unattainable, we regard the averaged marginal contributions from 10k randomly sampled orderings as the SHAP values for comparison (see supplementary materials for the justification of this assumption)
\subsection{Approximation performance of LIFT-CAM}\label{sec:4.2}
In this section, we quantitatively assess how precisely LIFT-CAM estimates $\alpha^{\text{shap}}$. Since the exact $\alpha^{\text{shap}}$ is unattainable, we regard $\alpha$ from SHAP-CAM$_{10\text{k}}$ as $\alpha^{\text{shap}}$ for comparison (see Supplementary Material for the justification of this assumption). Table \ref{table:cos_sim} shows the cosine similarities between $\alpha$ from state-of-the-art CAMs, including LIFT-CAM, and $\alpha$ from SHAP-CAM$_{10\text{k}}$.\par
As shown in Table \ref{table:cos_sim}, LIFT-CAM presents the highest similarities for all datasets (greater than 0.9), which indicates high relevance between $\alpha^{\text{lift}}$ and $\alpha^{\text{shap}}$. Even if Ablation-CAM also exhibits high similarities, the method falls behind LIFT-CAM due to dissatisfaction of the local accuracy. The other CAMs cannot approximate $\alpha^{\text{shap}}$, providing consistently low similarities.

\begin{table}[t]
\begin{center}
\begin{tabular}{lccc}
\toprule
     & \emph{ImageNet} & \emph{VOC} & \emph{COCO} \\
\midrule
     Grad-CAM & 0.489 & 0.404 & 0.441\\ 
     Grad-CAM++ & 0.385 & 0.329 & 0.412\\
     XGrad-CAM & 0.406 & 0.327 & 0.350\\
     Score-CAM & 0.195 & 0.181 & 0.157\\
     Ablation-CAM & 0.972 & 0.888 & 0.908\\
     %LRP-CAM & 0.969 & 0.865 & 0.877 \\
     LIFT-CAM & \underline{\textbf{0.980}} & \underline{\textbf{0.918}} & \underline{\textbf{0.924}}\\
\bottomrule
\end{tabular}
\end{center}
\caption{Cosine similarities between the coefficients from various CAMs and those from SHAP-CAM$_{10\text{k}}$. We analyze 500 randomly selected images for each dataset.}
\label{table:cos_sim}
\end{table}

\begin{figure*}[t]
\begin{center}
    \includegraphics[width=6.2in]{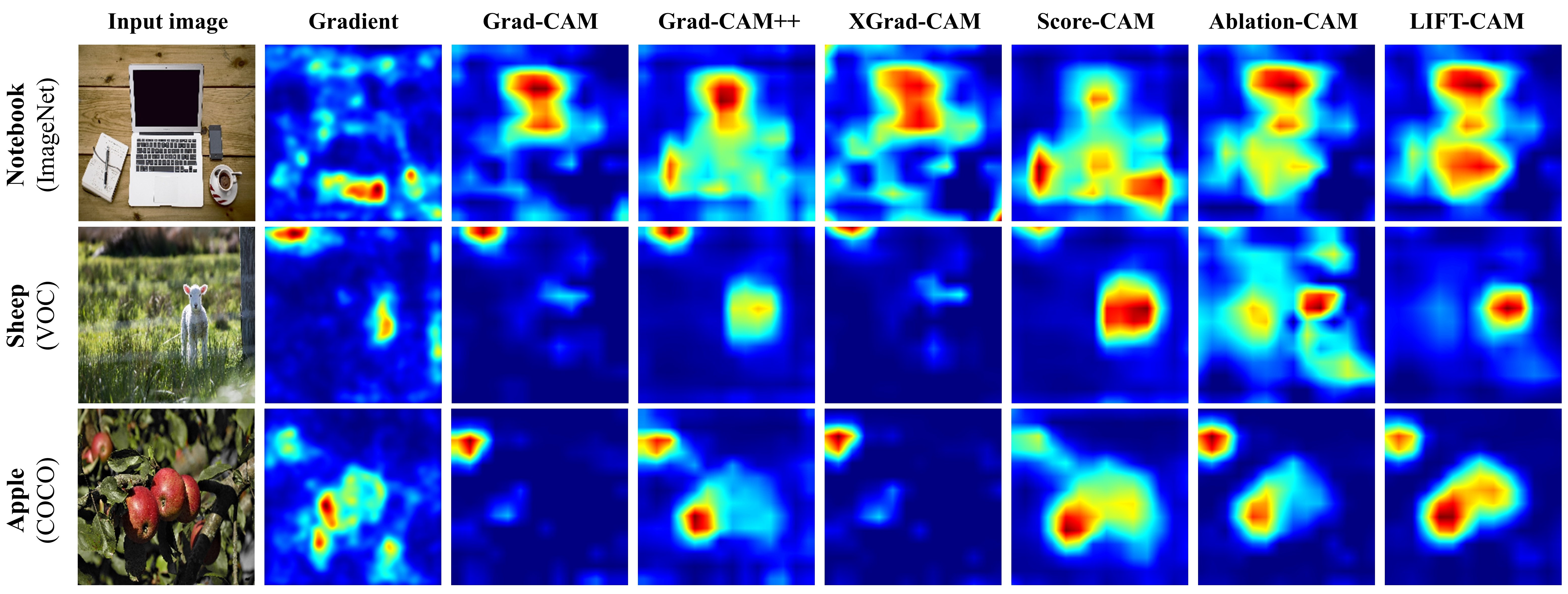}
\end{center}
   \caption{Visual explanation maps of state-of-the-art saliency methods and our proposed LIFT-CAM. Note that we apply a smoothing technique to Gradient \cite{grad} to acquire comparable visual explanation maps to those of CAMs, referring to \cite{fong2019understanding}.}
\label{fig:qualitative_results}
\end{figure*}

\begin{table*}[h]
\begin{center}
\begin{tabular}{lccccccccc}
\toprule
     & \multicolumn{3}{c}{Increase in Confidence (\%)} & \multicolumn{3}{c}{Average Drop (\%)} & \multicolumn{3}{c}{Average Drop in Deletion (\%)} \\ 
\cmidrule(lr){2-4} \cmidrule(lr){5-7} \cmidrule(lr){8-10}
    & \emph{ImageNet} & \emph{VOC} & \emph{COCO} & \emph{ImageNet} & \emph{VOC} & \emph{COCO} & \emph{ImageNet} & \emph{VOC} & \emph{COCO} \\
\midrule
    Grad-CAM & 24.0 & 32.7 & 31.9 & 31.89 & 30.73 & 30.74 & 30.60 & 17.43 & 25.66\\
    Grad-CAM++ & 23.1 & 33.8 & 33.5 & 30.53 & 17.20 & 20.87 & 27.98 & 15.85 & 24.16\\
    XGrad-CAM & 25.0 & 30.5 & 31.3 & 31.36 & 30.04 & 29.92 & 30.48 & 17.09 & 24.95 \\
    Score-CAM & 22.8 & 29.4 & 23.9 & 29.91 & 17.49 & 23.66 & 27.52 & 14.12 & 17.35 \\
    Ablation-CAM & 24.1 & 34.4 & 35.0 & 29.41 & 25.49 & 23.99 & 32.52 & 19.42 & \underline{\textbf{26.75}}\\
    %LRP-CAM & 25.1 & 34.3 & 33.6 & 28.56 & 25.18 & 26.49 & 32.91 & 18.79 & 26.01 \\
    LIFT-CAM & \underline{\textbf{25.2}} & \underline{\textbf{38.7}} & \underline{\textbf{39.3}} & \underline{\textbf{29.15}} & \underline{\textbf{17.15}} & \underline{\textbf{18.65}} & \underline{\textbf{32.95}} &  \underline{\textbf{20.09}} & 26.34 \\
\bottomrule
\end{tabular}
\end{center}
\caption{Comparative evaluation of faithfulness on the object recognition task between various CAMs. We analyze 1,000 randomly selected images for each dataset (the same image samples as Table \ref{table:proof_shap}). Higher is better for the IC and ADD. Lower is better for the AD.}
\label{table:comp_cam}
\end{table*}
%Higher is better for the increase in confidence and average drop in deletion. Lower is better for the average drop.

\subsection{Performance evaluation of LIFT-CAM}\label{sec:4.3}
The experimental results from Secs. \ref{sec:4.1} and \ref{sec:4.2} motivate us to generate visual explanations using LIFT-CAM. 
We verify the effectiveness of our LIFT-CAM by comparing the performances of the method with those of previous state-of-the-art saliency methods in terms of the quality of visualization, faithfulness, and localization.

\subsubsection{Qualitative evaluation}
Figure \ref{fig:qualitative_results} provides qualitative comparisons between various saliency methods via visualization. Each row represents the visual explanation maps for each dataset. When compared to the other methods, our proposed method, LIFT-CAM, yields visually interpretable explanation maps for all cases. It clearly pinpoints the essential parts of the specific objects which are responsible for the classification results. This can be observed in the notebook case (row 1), for which the other visualizations cannot decipher the lower part of the notebook. Furthermore, LIFT-CAM alleviates pixel noise without highlighting trivial evidence. In the sheep case (row 2), the artifacts of the image are eliminated by LIFT-CAM and the exact location of the sheep is captured by the method. Last, the method successfully locates multiple objects in the apple case (row 3) by providing the clear object-focused map.\par

\subsubsection{Faithfulness evaluation}
\noindent \textbf{IC, AD, and ADD.} Table \ref{table:comp_cam} shows the results of the IC, AD, and ADD for various CAMs. The three metrics can represent the object recognition performances of the saliency methods in a complementary way. LIFT-CAM provides the best results in most cases, with an exception of the ADD in COCO, where Ablation-CAM outperforms LIFT-CAM. However, LIFT-CAM also provides a comparable result, showing that the difference is negligible. In addition, it should be noted that LIFT-CAM is much faster than Ablation-CAM since it requires only a single backward pass to calculate the coefficients. Thus, LIFT-CAM can determine which object is responsible for the model's prediction, accurately and efficiently. \par

\noindent \textbf{Area under probability curve.} The above three metrics tend to be advantageous for methods which provide explanation maps of large magnitude. To exclude the influence of the magnitude, we can binarize the explanation map with two opposite perspectives: \textit{insertion} and \textit{deletion} \cite{rise}. First, we threshold $s(u(L^c_{\text{CAM}}))$ with $\delta \in [0:0.025:1]$ (i.e., 1 for top $100 \times \delta$ percentile pixels and 0 for the others) and acquire the corresponding target softmax outputs $O^c$ for the insertion and $D^c$ for the deletion. Using the softmax outputs, we can draw a probability curve as a function of $\delta$. Finally, we can calculate the area under the probability curve (AUC). \par
As shown in Table \ref{table:auc}, LIFT-CAM provides the most reliable results, presenting the highest insertion AUC and the lowest deletion AUC. Through this experiment, we demonstrate that LIFT-CAM succeeds in sorting the pixels according to the contributions to the target result.

\begin{table}[t]
\begin{center}
\begin{tabular}{lcc}
\toprule
     & Insertion & Deletion \\
\midrule
     Grad-CAM & 0.4427 & 0.0891\\
     Grad-CAM++ & 0.4350 & 0.0969\\
     XGrad-CAM &  0.4457 & 0.0883\\
     Score-CAM & 0.4345 & 0.1002\\
     Ablation-CAM & 0.4685 & 0.0873\\
     %LRP-CAM & 0.4682 & 0.0871\\
     LIFT-CAM & \underline{\textbf{0.4712}} & \underline{\textbf{0.0866}}\\
\bottomrule
\end{tabular}
\end{center}
\caption{AUC results in terms of the insertion and deletion. The values are averaged for 1,000 randomly selected images from ImageNet. Higher is better for the insertion and lower is better for the deletion.}
\label{table:auc}
\end{table}
% Higher is better for insertion and lower is better for deletion.

\begin{table*}[t]
\begin{center}
\begin{tabular}{lccccccc}
\toprule
     & Grad-CAM & Grad-CAM++ & XGrad-CAM & Score-CAM & Ablation-CAM & LIFT-CAM \\
\midrule
     Proportion (\%) & 47.76 & 49.14 & 47.91 & 51.14 & 51.87 & \underline{\textbf{52.43}}\\
\bottomrule
\end{tabular}
\end{center}
\caption{Proportions of energy located in bounding boxes for various CAMs. The values are averaged for 1,000 randomly selected images from ImageNet. Higher is better.}
\label{table:epg}
\end{table*}
%LRP-51.80

\subsubsection{Localization evaluation}
It is reasonable to expect that a dependable explanation map overlaps with a target object. Therefore, we can also assess the reliability of the map via localization ability in addition to the softmax probability. \cite{score_cam} newly proposed an improved version of a pointing game \cite{pointing_game}, named as an energy-based pointing game. This gauges how much energy of the explanation map interacts with the bounding box of the target object. For this, an evaluation metric can be formulated as follows:
\begin{equation}\label{eqn:epg}
    \text{Proportion} = \frac{\sum_{(i,j)\in bbox} s(u(L^c_{\text{CAM}}))_{(i,j)}}{\sum_{(i,j)\in \Lambda^{'}} s(u(L^c_{\text{CAM}}))_{(i,j)}}
\end{equation}
where $\Lambda^{'} = \{1, \dots, H^{'}\} \times \{1, \dots, W^{'}\}$ is an original image dimension and $s(u(L^c_{\text{CAM}}))_{(i,j)}$ denotes a min-max normalized importance at pixel location $(i,j)$. Higher is better for this metric. \par
The proportions of various methods are reported in Table \ref{table:epg}. LIFT-CAM shows the highest proportion compared to the other methods. This implies that LIFT-CAM produces a compact explanation map which focuses on the essential parts of the images without trivial noise.

\subsection{Application to VQA}\label{sec:4.4}
We also apply LIFT-CAM to VQA to demonstrate the applicability of the method. We consider the standard VQA model \cite{vqa} which consists of a CNN and a recurrent neural network in parallel. They function to embed images and questions, respectively. The two embedded vectors are fused and entered into a classifier to produce an answer. \par
\begin{figure}[t]
\begin{center}
   \includegraphics[width=0.95\linewidth]{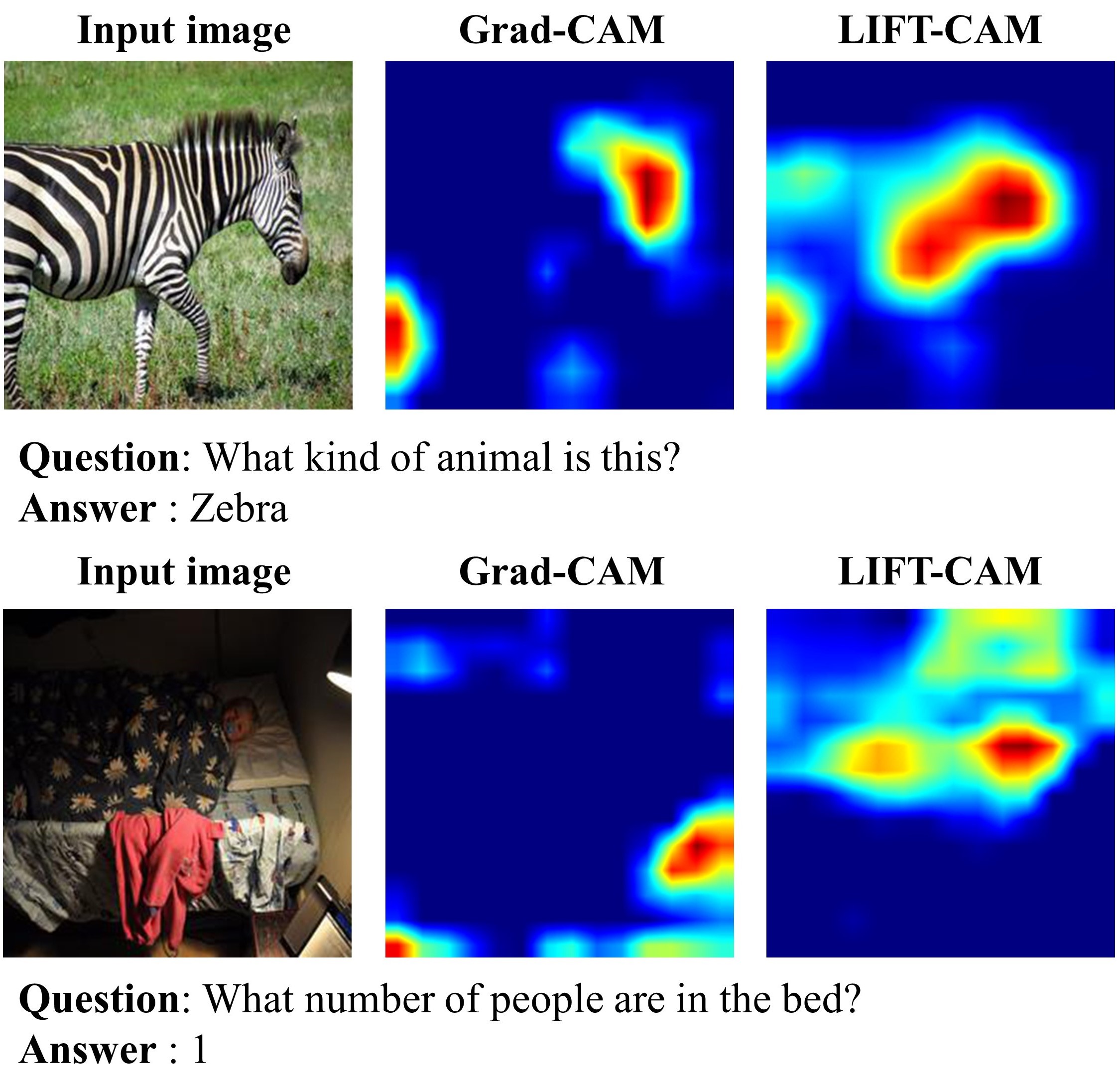}
\end{center}
   \caption{Visual explanation maps of Grad-CAM and LIFT-CAM for VQA.}
\label{fig:vqa}
\end{figure}

Figure \ref{fig:vqa} illustrates the explanation maps of Grad-CAM and LIFT-CAM for VQA. LIFT-CAM highlights the regions in the given images that are more relevant to the question and answer pairs than those identified with Grad-CAM. Additionally, since this is a classification problem, the IC, AD, and ADD can be evaluated with fixed question embeddings. Table \ref{table:vqa} shows the comparison results between Grad-CAM and LIFT-CAM in terms of those metrics. As demonstrated in the table, LIFT-CAM outperforms Grad-CAM for all of the metrics. This indicates that LIFT-CAM is better at figuring out the essential parts of images, which can serve as evidences for the answers to the questions.

\begin{table}[t]
\begin{center}
\begin{tabular}{lcc}
\toprule
     & Grad-CAM & LIFT-CAM \\
\midrule
     Increase in Confidence (\%) & 41.95 & \underline{\textbf{43.39}}\\
     Average Drop (\%) & 16.71 & \underline{\textbf{14.14}}\\
     Average Drop in Deletion (\%) & 9.09 & \underline{\textbf{9.58}}\\
\bottomrule
\end{tabular}
\end{center}
\caption{Faithfulness evaluation for the VQA task. The values are averaged for the complete validation set. Higher is better for the IC and ADD. Lower is better for the AD.}
\label{table:vqa}
\end{table}
% Higher is better for the increase in confidence and average drop in deletion. Lower is better for the average drop.

\section{Conclusion}
In this work, we propose a novel analytic framework which determines the coefficients of CAM by optimizing a linear explanation model, using additive feature attribution methods. As desirable solutions for the explanation model, we introduce several approaches including LIFT-CAM, throughout this paper and Supplementary Material. In addition, we show that Ablation-CAM can also be unified into this framework. \par
Our proposed LIFT-CAM approximates the SHAP values of the activation maps, which is a unified solution for the explanation model, with a single backward pass. The method provides qualitatively enhanced visual explanations compared with the other CAMs. Furthermore, it achieves state-of-the-art results on various quantitative evaluation metrics.
{\small
\bibliographystyle{ieee_fullname}
\bibliography{ref}
}

\end{document}